# Fitting New Speakers Based on a Short Untranscribed Sample

Eliya Nachmani [1]  Adam Polyak [1]  Yaniv Taigman [1]  Lior Wolf [1,2]

## Abstract

Learning-based Text To Speech systems have the potential to generalize from one speaker to the next and thus require a relatively short sample of any new voice. However, this promise is currently largely unrealized. We present a method that is designed to capture a new speaker from a short untranscribed audio sample. This is done by employing an additional network that given an audio sample, places the speaker in the embedding space. This network is trained as part of the speech synthesis system using various consistency losses. Our results demonstrate a greatly improved performance on both the dataset speakers, and, more importantly, when fitting new voices, even from very short samples.

## 1. Introduction

The technological feasibility of ubiquitous Text To Speech (TTS), in which talking avatars of everyone we know would interact with us as a form of asynchronous communication, depends on the ability to sample individual speakers in a casual way. In the past year, Neural TTS systems have shifted from high quality single speaker systems to multi-speaker systems (see Sec. 2). However, most of these systems rely on the availability of training samples of all speakers during training.

One exception is the VoiceLoop system (Taigman et al., 2018), which was shown to be able to fit a new speaker from relatively few samples. This ability, however, is limited by three factors: (i) The obtained quality of a training-naive voice is lower than the quality obtained for speakers that participated in the training process. (ii) The ability of the system to capture the identity of a new speaker, heavily depends on the amount of data that is available for the fitting process. (iii) The fitting requires both the voice sample and the transcript.

In this work, we propose a TTS network that is designed to fit a new voice based on a limited amount of data and without the transcript of the new speaker. While the current multi-speaker TTS methods rely on a speaker embedding that is stored in one or more Look Up Tables (LUTs), our method incorporates a fitting network $N_s$, which is trained jointly with the other networks of the TTS system.

During training, instead of retrieving the speaker embedding for the speaker of the training sample from the LUT, we apply the network $N_s$ to the sample's target audio. The embedding thus obtained is used in the process of generating the same audio. Multiple losses are added in order to ensure that the embedding of the reconstructed audio is similar to the embedding of the target sample and that the embeddings are well-separated.

While our approach is general and can be applied to any neural TTS system, we focus our experiments on the VoiceLoop system, since it is the only multispeaker system for which an official implementation was released. The official implementation of Char2Wav (Sotelo et al., 2017) is multi-speaker, but the method was published as a single speaker method and is not competitive with other multispeaker systems. DeepVoice3 has a community implementation (Park, 2018). However, at this time, the generated voice contains noticeable artifacts.

In addition to proposing the new method, we also investigate the underlying principles of voice fitting by constructing a single network that does not model individual speakers, but which is made to mimic multiple speakers by performing priming. These experiments demonstrate that mimicking a new speaker does not require an optimization process and can be based on even a short sample.

## 2. Previous Work

A common view of the TTS literature divides the existing methods into four top-level groups: rule-based methods, concatenative systems, statistical-parametric, which includes many successful HMM based methods, and the emerging neural models.

While there have been attempts to create concatenative systems that rely on relatively little data (Jin et al., 2017), or on automatic filtering of the training set used for an HMM system (Baljekar & Black), concatenative and statistical-parametric systems still require clean and well transcribed samples. These samples are of a minimal length of tens of minutes and sometimes require the reading of specific sentences in order to cover the entire space of combinations of consecutive phonemes.

---

[1]Facebook AI Research [2]Tel Aviv University. Correspondence to: Eliya Nachmani <eliyan@fb.com>.

The situation is not yet materially different for neural speech systems. A critical look at the current neural methods reveals that the most striking voice qualities are obtained on single speaker models, trained on hours of carefully transcribed samples and that the literature on multispeaker systems does not focus on de-novo speakers.

The recent neural TTS systems include the Deep Voice systems DV1 (Arik et al., 2017b), DV2 (Arik et al., 2017a) and DV3 (Ping et al., 2018), WaveNet (Oord et al., 2016) and Parallel WaveNet (van den Oord et al., 2017), Char2Wav (Sotelo et al., 2017), Tacotron (Wang et al., 2017) and Tacotron2 (Shen et al., 2017), and VoiceLoop (Taigman et al., 2018). One can sort these systems according to several axes, which are of relevance to our work: the input of the system, the output of the system, the underlying architecture, the ability to work with multiple speakers, the nature of the embedding of multiple speakers, and the ability to fit new speakers.

The literature contains three types of features, with some systems tested on multiple types: (i) raw letters (Tacotron, Tacotron2, DV3), (ii) phonemes (Char2Wav, VoiceLoop, DV3), and (iii) linguistic features, including duration and pitch of phonemes (WaveNet, Parallel WaveNet, DV1, DV2). The latter requires a dedicated system to extract these features.

With regards to output, there are also a few options in the literature: Tacotron creates spectrograms, which are inverted using the Griffin-Lim method. Char2Wav and Voice Loop produce World Vocoder Features (Morise et al., 2016). The WaveNet and Parallel WaveNet systems create raw audio. DV1, DV2, DV3 and Tacotron2 employ WaveNet as neural vocoder to transform compact representations, such as spectrograms or vocoder features to raw audio.

There is a variability in the literature also for the underlying method. Wavenet employs dilated convolutions. The Tacotron method employs multiple RNNs, convolutions and a highway network (Srivastava et al., 2015). Recently, Tacotron2 simplified the latter by replacing highway networks with RNNs and predicting a residual to improve the system output. DV1 and DV2 employ bidirectional RNNs, multilayer fully connected networks and residual connections. DV3 simplified the previous systems by using a convolutional sequence to sequence architecture (Gehring et al., 2017) and incorporating the key-value attention mechanism of (Vaswani et al., 2017). Finally, the VoiceLoop method is based on a shifting buffer, which is updated in a FIFO manner.

In all three categories above, we follow VoiceLoop: our method relies on phonemes, produces world vocoder features, and shares large parts of our architecture with it, including the shifting buffer-based RNN. The reliance on World Vocoder is convenient, especially considering the landscape of open-source WaveNet implementations with regards to quality and efficiency. However, it upper-bounds the obtained quality.

Only three published neural systems are multi-speaker: DV2, DV3, and VoiceLoop[1]. All three systems were tested on the VCTK dataset, which contains over a hundred speakers. DV2 was also applied to an internal dataset of audiobooks with 447 speakers and DV3 to LibriSpeech (Panayotov et al., 2015) with 2484 speakers. The VoiceLoop system was demonstrated on a dataset of four "in-the-wild" speakers collected from YouTube videos of public speeches. Our method is evaluated, among other datasets, on the VoxCeleb dataset (Nagrani et al., 2017), which is of lower quality and more uncontrolled and heterogeneous than any other existing corpus.

As mentioned, out of these systems, the only one that was demonstrated to fit new speakers, which were not encountered during training, is the VoiceLoop system. Unlike our system, this is done by an optimization process in which the embedding of a new speaker is searched by repeating the training process for her samples, while fixing all weights, except for the embedding of that speaker. Therefore, a lengthy backpropagation fitting phase is needed in order to obtain text to speech in a new voice. We show that this process is both unnecessary and leads to suboptimal performance.

### 2.1. The VoiceLoop Architecture

At the heart of VoiceLoop is a buffer $S_t$, which serves as a differentiable memory and is used by all of the model's networks. At every time step, a new representation vector $u_t$ is inserted into the buffer, and the first inserted vector is discarded.

The VoiceLoop model is composed out of three shallow fully connected networks ($N_u$, $N_a$, $N_o$), two LUTs ($LUT_p$, $LUT_s$), and two projection matrices ($F_u$ and $F_o$). $N_u$ creates the new vector $u_t$, $N_a$ updates the attention mechanism, and $N_o$ generates the next audio frame, $o_t$, which is encoded as vocoder features. The input to the VoiceLoop system is a sequence of phonemes $s_1, s_2, \ldots, s_l$, which are converted to a sequence of embedding vectors, given by $LUT_p$ and stored as the columns of a matrix $E$. $LUT_s$ and the projection matrices allow the multi-speaker behavior, and each speaker is represented by an embedding vector $z$, which is

---

[1] Wavenet was shown to produce mumbling of multiple speakers, but was not demonstrated as a multi-speaker TTS system

stored in this LUT.

Each forward pass runs three sequential steps. In the first step, the context vector is computed. The Graves monotonic attention mechanism (Graves, 2013) is used: the attention network $N_a$ receives the current buffer and outputs the priors of the Gaussian Mixture Model, shifts of the means of the Gaussians, and their log-variances. The current context vector $c_t$ is then computed as a weighted sum of the columns of the input sequence embedding matrix $E$.

In the next step, a representation vector $u_t$ is added to the buffer at the first location $S_t[1]$, shifting all other buffer locations to the right: $S_t[i+1] = S_{t-1}[i]$ for $i = 1, \ldots, k-1$. $u_t$ is computed by $N_u$ using the buffer $S_{t-1}$, the sum of the context vector $c_t$ and the projection of the speakers embedding $F_u z$, and the previous output $o_{t-1}$.

This output vector $o_t$ is generated by the third and last step by the network $N_o$, whose inputs are the buffer $S_t$ and the projection of the user by $F_o$.

Training is done by minimizing the MSE of the output vocoder features. Since this procedure assumes that the generated output is perfectly aligned with the target, teacher-forcing is used. In other words, during training, the network receives the correct output of the previous frame in lieu of the predicted target $o_{t-1}$.

In this work, we remove the speaker-embedding LUT ($LUT_s$) and employ an additional network that transforms an input audio clip to the representation $z$. By doing so, we are able to have the task of fitting as part of the method and not as an afterthought. This enables efficient fitting from a short sample and without the transcribed text.

## 3. The Voice Constancy Phenomenon

The neural multispeaker systems in the literature (Arik et al., 2017a; Ping et al., 2018; Taigman et al., 2018) are based on embedding the speaker's voice in some vector space. This form of embedding opens the way to fitting a new speaker, using backpropagation, as was shown in (Taigman et al., 2018).

In this section, we aim to show that speaker embedding is not the only possible way to construct a multi-speaker system and that the fitting process does not require an optimization step. While the technique we present here is detached from fitting method that is the focus of our paper, it provides important insights on the behavior of multi-speaker voice generating systems as well as on the length of data required in order to capture the voice of new speakers. Namely, it shows that a single network can be trained for all speakers in the training set, without conditioning on the speaker, and that a new speaker can be captured from a very short sample.

This is demonstrated by training a single network, in which all identity related elements are removed. i.e., We remove the speaker embedding $LUT_s$ and the two projection matrices $F_u$ and $F_o$. While training, we make no use of the identity information in any way, creating a network that is agnostic to the speaker.

In order to generate a specific voice, we make use of the property that we call voice-constancy. Namely, that a network that generates a sample in a certain voice would continue to employ that voice in subsequent frames. This property is an outcome of the training process, in which the input regarding the speaker is not given to the network and is only evident as part of the samples of the previous time steps, provided during training, due to the teacher-forcing procedure.

In order to speak in a certain voice we, therefore, use priming (Graves et al., 2014). We play a short sample, typically of 300 frames (1500 milliseconds), using a teacher forcing procedure. We then let the generation process continue without resetting the buffer. In other words, we continue the voice generation using the primed buffer and the new input text. The results obtained using the priming-based approach are presented in Sec. 5.

## 4. The Fitting Sub-Network

The priming-based method presented in the previous section is simple to implement and moderately effective. However, we were not able to make it work better than the baseline VoiceLoop system. In addition, it requires, during fitting, the textual transcription as well as the audio sample. While an automatic speech recognition system can be used to extract this transcript, it is still limiting for several reasons: (i) speech recognition (i.e., speech to text) systems do not exist for most of the world's languages. (ii) the accuracy of these systems is limited, especially in uncontrolled settings. (iii) for the existing speech recognition systems, aligning the produced transcript to the audio, as well as extracting the phonemes requires addition preprocessing steps.

We modify the VoiceLoop system, described in Sec. 2.1, by incorporating a fitting network $N_s$, which given an audio sample $\mathbf{y} = y_1, y_2, \ldots, y_m$ produces an embedding vector $z$. This embedding vector is then used in the VoiceLoop networks as the speaker's embedding.

$N_s$ receives as input a tensor of size $1 \times m \times d_o$, which is length of the audio times the size of the vocoder feature vector, set to 63 for the World Vocoder features (Morise et al., 2016). The network has five convolutional layers of $3 \times 3$ filters, each with 32 channels. Batch normalization is performed after each convolutional layer, followed by a ReLU activation function. Following the convolutional layers, average pooling over time is performed, followed by two fully-connected layers, of size 256 each, with ReLU activations. Finally, an affine projection followed by an $L_2$ normalization is performed in order to obtain the embedding vector $z$.

### 4.1. The Loss Term

The VoiceLoop system is trained using the MSE loss, given an audio sample $\mathbf{y}$:

$$L_{\text{MSE}} = \frac{1}{d_o} \sum_{\mathbf{y}} \sum_{t=1}^{l} \|y_t - o_t\|^2 \quad (1)$$

where both the ground truth $y_t$ and the network's output $o_t$ are vectors in $\mathbb{R}^{d_o}$. Since teacher forcing is used, the two sequences are of the same length. Note that in our method, $o_t$ is both a function of $\mathbf{y}$, via $N_s$ and of a sequence of phonemes $\mathbf{s}$, as depicted in Fig. 1.

We add two additional losses. Given three voice samples: $\mathbf{y}^1, \mathbf{y}^2, \mathbf{y}^3$ such that $\mathbf{y}^1, \mathbf{y}^2$ are from the same speaker, and $\mathbf{y}^3$ is not, we would like the computed embedding of the first two samples to be similar to each other, while different from that of the third. A contrastive loss term with a margin $\Delta$ is used:

$$L_{\text{contrast}} = \frac{1}{2} \sum_{\mathbf{y}^1, \mathbf{y}^2, \mathbf{y}^3} (\|N_s(\mathbf{y}^1) - N_s(\mathbf{y}^2)\|^2 \\ + \max(0, \Delta - \|N_s(\mathbf{y}^2) - N_s(\mathbf{y}^3)\|)^2), \quad (2)$$

where the sampling of a triplet is done by organizing each batch, such that it contains a sequence of pair of samples from the same speaker, and joining the first sample of the next pair to each pair. In all our experiments, we use a margin of $\Delta = 1$.

In addition, in order to require speaker constancy between the input audio clip and the generated audio clip, we add a third loss term. Given an input audio $\mathbf{y}$, we compute the embedding using $N_s$, run VoiceLoop with this embedding, obtaining an output audio $\mathbf{o} = o_1, o_2, \ldots, o_l$, to which we apply $N_s$ again. The loss term is then defined as:

$$L_{\text{cycle}} = \sum_{\mathbf{y}} \|N_s(\mathbf{y}) - N_s(\mathbf{o})\|^2 \quad (3)$$

The overall loss is a weighted combination of the three losses:
$$L = L_{\text{MSE}} + \alpha L_{\text{contrast}} + \beta L_{\text{cycle}}, \quad (4)$$

where in all of our experiments, we set $\alpha = \beta = 10$.

### 4.2. Training Details

Similar to Taigman et al. (2018), we train our networks in two phases. The first phase employs data with larger amounts of added i.i.d white noise, while the subsequent phase is trained on longer sequences to which less noise was applied. Specifically, during the first phase, a noise SD equal to 4.0 is added to the sequences of ground truth vocoder features ($\mathbf{y}$) and these sequence are cropped to a length of 100. A batch size equal to 256 is used for exactly 90 epochs. Phase 2 of the training process employs noise SD of 2.0, and sequence lengths that are trimmed at 1000 vocoder features. The batch size is reduced to 30, in order to fit longer sequences in memory. This phase is run until convergence.

The various parameters follow the implementation released by Taigman et al. (2018). A few modifications are done to the attention mechanism. First, a $\tanh$ activation function is applied to the output of the network $N_a$. Second, we found that a slight improvement is obtained if, during inference only, the mixture component with the maximal prior is selected, instead of employing the weighted sum to obtain the next shift in the attention position.

We also added a balanced mini-batch mechanism to speed up the training iterations. This is done by splitting the training dataset into four equally sized partitions based on the audio sample length and sampling each mini-batch from one of the splits. As a result, each sample in a specific mini-batch requires less padding, which in-turn reduces the total amount of computation required from the model.

## 5. Experiments

We focus on multispeaker TTS and especially on fitting. Since VoiceLoop is the only open implementation of a multispeaker contribution, and since it is the only contribution to demonstrate fitting to new speakers, we employ it as our baseline. We also evaluate the priming based method of Sec. 3 and also perform an ablation analysis to demonstrate the contribution of the various components of our loss.

**Evaluation Metrics** A comparison between generative methods in a non-deterministic setting is always a challenge. However, following the literature, we employ a sufficient number of tools in order to demonstrate the gap in performance. In order to evaluate the quality of the generated audio, we employ both the Mean Opinion Scores (MOS) and the Mel Cepstral Distortion (MCD) scores. To evaluate the identifiability of the generated voice, we employ either multiclass classification on a network trained on the VCTK identities, or the ROC statistics obtained for the same/not-same task using the embedding layer of the same speaker identification network.

The MOS measure is obtained using the crowdMOS toolkit by (P. Ribeiro et al., 2011) and Amazon Mechanical Turk (AMT). The samples were presented at a fixed framerate of 16kHz and at least 20 raters participated in each such experiment, with 95% confidence intervals. All AMT experiments were restricted to North American raters.

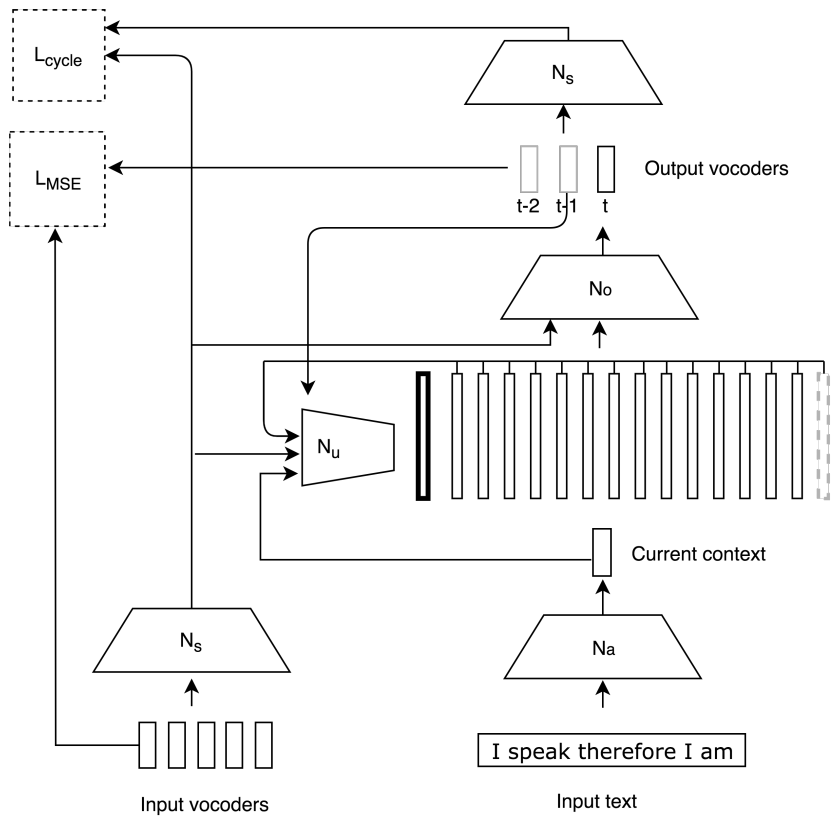

*Figure 1.* **Model Architecture.** The trapezoids denote the various networks, and the dashed boxes the losses. $L_{\text{contrast}}$ is not shown, since it involves multiple samples.

MCD is an automatic method of testing compatibility between the spectra of two audio sequences, which is limited to specific aspects of the quality. Since the sequences generated are not aligned, the MCD DTW is used, in which Dynamic Time Warping aligns the sequences prior to comparison.

Beyond quality, the generated voices need to comply with the target voice. Following (Arik et al., 2017a; Taigman et al., 2018), this is evaluated using a speaker recognition network. We train offline a network with the same architecture of $N_s$ for this purpose on the ground-truth training set of the VCTK speakers. For VCTK experiments, the network is tested on the generated sequences and the obtained accuracy is reported.

For datasets other than VCTK, we consider pairs of samples and compute distances using this identification network. Specifically, for each pair, we compute the cosine distance between the activations of the last layer prior to the classification, which is of dimensionality 256. As "same" pairs, we collect one previously unused real-voice sample of speaker A and one generated sample, using a voice that was fitted on another sample of speaker A. For the "not-same" pair, the process is identical, except that the second sample is collected from speaker B. We then compute the ROC curve, using all same pairs and a large sample of not-same pairs, and report the Area Under Curve (AUC).

**Datasets** The VCTK dataset (Veaux et al., 2017) contains 109 speakers. In order to make our experiments compatible with the results reported by (Taigman et al., 2018), we employ their subset of 85 speakers for training our network and another subset of 16 speakers for the fitting experiments. The remaining eight speakers, which were left out for validation, are not used in our experiments. Among the samples of each speaker, we use the existing splits of train and test.

The LibriSpeech dataset (Panayotov et al., 2015) is a corpus of 360 hours of voice that was compiled out of audio books from the LibriVox collection of free public domain audiobooks. Due to the expected training time, we focus in this submission on a subset we call "Libri-15GB", which is comprised of the first 15GB of the dataset, when sorting the speakers alphabetically . The fitting experiments are done on speakers from the rest of the dataset ("Libri-rest").

Table 1. MOS of trained voices (Mean ± SE)

| Method | VCTK85 | Libri15GB |
|---|---|---|
| VoiceLoop | 3.33 ± 1.00 | 2.19 ± 1.16 |
| Our | 3.52 ± 0.88 | 2.31 ± 1.13 |
| Ground truth | 4.67 ± 0.62 | 4.60 ± 0.70 |

Table 2. MCD scores (lower is better) of trained voices (Mean ± SE)

| Method | VCTK85 | Libri15GB |
|---|---|---|
| VoiceLoop | 14.16 ± 0.87 | 9.92 ± 1.94 |
| Our | 13.90 ± 0.85 | 8.80 ± 1.00 |

Table 3. Top-1 identification accuracy (VCTK85) and AUC (Libri15GB) for the trained voices.

| Method | VCTK85 Accuracy | Libri15GB AUC |
|---|---|---|
| VoiceLoop | 100% | 0.84 |
| Our | 99.49% | 0.89 |
| Ground truth | 98.25% | 0.93 |

Table 4. MOS of fitted voices (Mean ± SE)

| Method | VCTK16 | Libri-rest | VoxCeleb |
|---|---|---|---|
| VoiceLoop | 2.98 ± 0.93 | – | – |
| Our | 3.66 ± 0.84 | 2.53 ± 1.11 | 2.16 ± 1.02 |
| Ground truth | 4.61 ± 0.68 | 4.43 ± 0.78 | 4.34 ± 0.88 |

The VoxCeleb dataset (Nagrani et al., 2017) is a compilation of YouTube urls and time stamps, which were obtained using an automatic pipeline, which consists of video-based active speaker identification and face verification. The dataset is collected for the task of identification based on voice, and the quality of many of the audio clips is not high. Our system does not need the transcript for fitting. However, the baseline VoiceLoop method does. For this purpose, we employ the automatic transcript by YouTube. This transcript is also used to cut the dataset into individual sentences.

### 5.1. Evaluation of Trained Voices

We first evaluate the quality of the trained model on the trained identities. While this is not the focus of this work, it is important to validate that our approach, which forgoes the speaker LUT and replaces it with a network trained on short voice clips, does not result in a degradation in performance.

Tab. 1 depicts the MOS values obtained for generated test samples of the VCTK85 dataset, as well as the Libri15GB subset. As can be seen, the quality obtained with our model is higher than that of VoiceLoop. Similar conclusions can be drawn from Tab. 2, which shows the MCD scores. Note that as reported also in (Ping et al., 2018), LibriSpeech results are lower than VCTK results. This probably stems from the added prosody in audiobooks and from the inability of the systems to model long term interactions within paragraphs due to the training procedure.

In addition, in Tab. 3, we evaluate the identification accuracy for the VCTK85 generated voices as well as the AUC obtained on the same/not-same identification experiment on Libri15GB. As shown, our method, despite not using an explicit per-speaker embedding during training, presents a similar level of identifiability to that of VoiceLoop.

### 5.2. Fitting Experiments

In the next set of experiments, we evaluate our ability to capture new voices, unseen during training. For both our method and the baseline VoiceLoop method, and across all fitted datasets, models which were trained on VCTK85 were used.

In Tab. 4, we present the MOS values obtained for the fitted voices. In this experiment, all of the training samples of each new speaker were used (of course, each new speaker was fitted individually). As can be seen, our method shows a significant gap over the baseline VoiceLoop method for the 16 new speakers in VCTK16. A similar gap is shown for VCTK16 in Tab. 6 for the automatic MCD and the identity classification accuracy.

We do not present MOS results for VoiceLoop model fitted on Libri-rest and VoxCeleb dataset since the quality of the generated samples was not suitable for MOS experimentation. However, We do include identification results for the fitted VoiceLoop results as baselines on LibriSpeech and VoxCeleb.

Identification results for LibriSpeech are shown in Fig. 2 as ROC curves. As can be seen, our method outperforms VoiceLoop. The identification results for the fitted voices of VoxCeleb is not high, even for the original voice clips, due to a high intra-speaker variability and low audio quality. We, therefore, report these on subsets of the speakers which are stratified by quality. Specifically, we measure the inter-sample distances for each speaker and compare it to the average inter-speaker distance. We then threshold and employ only classes with a ratio (of the latter over the former) higher than a parameter. In Fig. 3, we present the AUC obtained in the same not same discrimination task as a function of this parameter. As can be seen, the identifiability obtained by our method is relatively high, when compar-

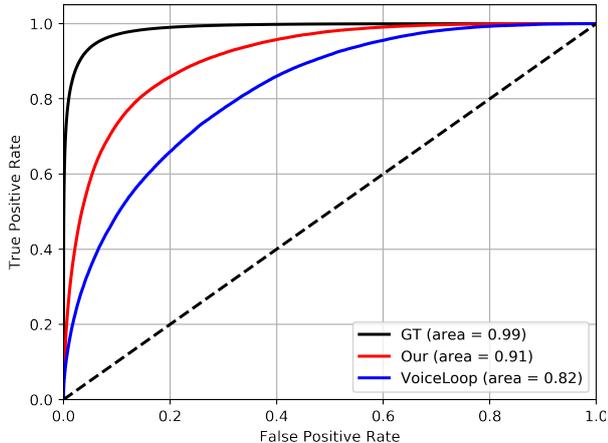

*Figure 2.* ROC for same/not-same identification on speakers from LibriSpeech. Ground truth (GT) is computed on pairs of real samples, while VoiceLoop and our method are measured on a real sample vs. a generated one.

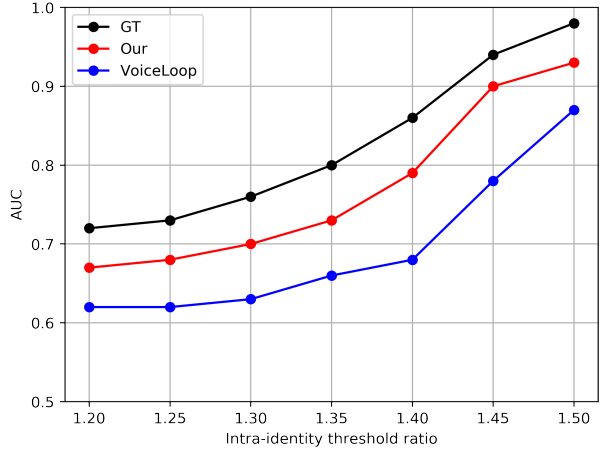

*Figure 3.* Identification ability on VoxCeleb as measured by AUC of the same/not-same task. Results are shown for the ground truth samples (GT), our method, and the VoiceLoop method as a function of a quality threshold applied to the speakers (see text).

ing to the score obtained by the ground truth samples. The AUC without filtering is 0.72 for the ground truth, 0.62 for VoiceLoop and 0.67 for our method

The ability to fit a new speaker is expected to depend on the length of the available sample. We, therefore, perform experiments exploring the quality of the fitted voices as a function of the maximal sample's length. Note that the average length is lower than the maximal length: if the length threshold is set at 15 min and there are only 12 min of that speaker, then only 12 min are used.

The results are presented in Tab. 5. As can be seen, our method is preferable across all lengths to VoiceLoop, in which the fitting procedure is a much lengthier process that involves backpropagation. Note that quality-wise (but not with regards to identification), the best results for our method are obtained when using a short sample of the speaker. This could be a consequence of the length of the sample used for $N_s$ during training, which, as detailed in Sec 4.2, is 0.5 sec in the first 90 epochs and up to 5 sec (but 3.1 sec on average) in the subsequent epochs. The Priming based method, presented in Sec. 3, shows relatively good quality but is not as identifiable.

### 5.3. Ablation Analysis

Using the automatic MCD and top-1 identification accuracy scores, we also compare with simplified versions of our method in which some of the losses are removed. Specifically, we compare with a version of our method in which $L_{\text{cycle}}$ is removed and another version in which $L_{\text{contrast}}$ is removed. As can be seen in Tab. 6 removing each of these terms leads to a significant loss of accuracy for the fitted voices, and in the case of $L_{\text{contrast}}$ also for the trained voices.

In order to further visualize this, Fig. 4 presents 2D TSNE plots for the embedding obtained from the VCTK16 samples with and without the $L_{\text{cycle}}$ term. Both these results are obtained without $L_{\text{contrast}}$, which pushes male and females so far that the TSNE plot is uninformative without zooming. As can be seen, $L_{\text{cycle}}$ contributes significantly to the separation of unseen voices in the embedding space.

### 5.4. Audio Samples

Various samples can be found on the project's webpage https://ytaigman.github.io/fitspk/index.html. These include generated samples of VCTK16 voices segregated by the length of the sample used for fitting and samples generated after fitting voices from both VoxCeleb and LibriSpeech-rest.

## 6. Conclusions

By demonstrating the ability to fit, in a feed-forward manner on even very short samples from uncontrolled ("in-the-wild") datasets, we brought neural TTS systems significantly closer to fulfilling their promise. The advancement we have made, although shown in the context of a specific architecture, is widely applicable and one can draw a few conclusions, which are unintuitive and even surprising.

First, identifiable voices can be captured from short sam-ples and without transcript. Machines are, therefore, able to mimic voices much more easily than what was previously believed. This is further demonstrated by presenting a single, simplified, network that given a second of speech can replicate its speaker approximately well. This capability is based on a priming operator, and relies on a phenomenon we identify, the voice constancy property.

Table 5. Quality scores for fitting the voices of VCTK16, for different maximal sample lengths (Mean ± SE)

| Metric | MOS | | | MCD | | | Top-1 accuracy | | |
|---|---|---|---|---|---|---|---|---|---|
| Method | VoiceLoop | Our | Priming | VoiceLoop | Our | Priming | VoiceLoop | Our | Priming |
| 1.5 sec | 2.92 ± 1.08 | 3.84 ± 0.92 | 3.30 ± 1.15 | 14.74 ± 1.06 | 14.35 ± 0.83 | 14.68 ± 0.75 | 52.55% | 73.72% | 22.02% |
| 1 sentence | 3.22 ± 1.09 | 3.94 ± 0.90 | – | 14.64 ± 0.97 | 14.35 ± 0.84 | – | 57.66% | 76.64% | – |
| 2 sentences | 2.94 ± 0.98 | 3.35 ± 1.14 | – | 14.72 ± 0.95 | 14.49 ± 0.83 | – | 65.14% | 77.52% | – |
| 1 min | 2.86 ± 0.99 | 3.47 ± 1.09 | – | 14.72 ± 0.97 | 14.47 ± 0.83 | – | 67.33% | 84.38% | – |
| 5 min | 3.07 ± 0.90 | 3.59 ± 1.06 | – | 14.61 ± 0.96 | 14.45 ± 0.83 | – | 77.19% | 84.38% | – |
| 10 min | 3.27 ± 0.95 | 3.86 ± 0.99 | – | 14.55 ± 0.96 | 14.46 ± 0.84 | – | 82.30% | 84.52% | – |
| 15 min | 3.14 ± 0.99 | 3.83 ± 1.04 | – | 14.53 ± 1.01 | 14.47 ± 0.83 | – | 83.58% | 84.38% | – |
| 20 min | 3.15 ± 0.96 | 3.74 ± 1.10 | – | 14.55 ± 0.98 | 14.45 ± 0.84 | – | 83.21% | 83.94% | – |

Table 6. MCD and identification accuracy in an ablation analysis performed on VCTK-85 and VCTK-16.

| Dataset | VCTK85 (Trained Voices) | | VCTK16 (Fitted Voices) | |
|---|---|---|---|---|
| | MCD | ACC | MCD | ACC |
| GT samples | | 98.25% | | 95.62% |
| VoiceLoop | 14.16 ± 0.87 | 100% | 14.72 ± 0.97 | 67.33% |
| Our | 13.90 ± 0.85 | 99.49% | 14.47 ± 0.83 | 84.38% |
| No $L_{\text{cycle}}$ | 14.07 ± 0.86 | 99.05% | 14.41 ± 0.79 | 77.37% |
| No $L_{\text{contrast}}$ | 14.07 ± 0.91 | 96.50% | 14.48 ± 0.80 | 56.20% |

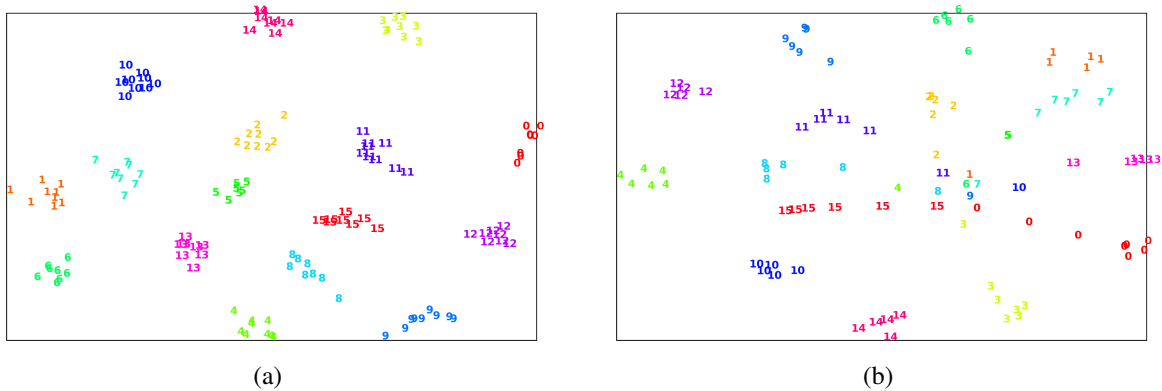

Figure 4. TSNE visualization of the learned embedding for VCTK16 with (a) and without (b) the loss term $L_{\text{cycle}}$. Each data point is the embedding, using $N_s$ of one voice sample. The numbers represent the various speakers, which are also color coded.

Second, by training on VCTK85 and then fitting on datasets with different characteristics and many more speakers, is becomes apparent that it is sufficient to train on a small population of 85 speakers in order to capture much of the variation in the general population.

Third, we demonstrate that a dynamic embedding, which is captured on-the-fly, is able to at least match learned embeddings. This is surprising, and as our ablation analysis shows, stems from the losses we incorporate into the problem.

One of the losses employed, $L_{\text{cycle}}$, opens the way for training a TTS system in a semi-supervised way, in which some speakers are transcribed and some are not. This is because it does not require that the audio generated by the system is identical to the input audio, only that the speaker identity is preserved. This way, the same TTS system can be trained on many languages at once, including languages without suitable transcribed corpora. This is left for future work.